\documentclass[conference]{IEEEtran}
\usepackage{times}

\usepackage[numbers]{natbib}
\usepackage{multicol}
\usepackage[bookmarks=true]{hyperref}
\usepackage{comment}
\usepackage{amsmath}

\DeclareMathOperator*{\argmax}{arg\,max}
\usepackage{times}
\usepackage{latexsym}
\usepackage{float}
\usepackage{booktabs}
\usepackage{graphicx}

\begin{document}

\title{Extracting Zero-shot Common Sense from Large Language Models for Robot 3D Scene Understanding}


\author{\authorblockN{William Chen, Siyi Hu, Rajat Talak, Luca Carlone}
\authorblockA{Laboratory for Information \& Decision Systems (LIDS)\\
Massachusetts Institute of Technology\\
Cambridge, Massachusetts 02139\\
\{verityw, siyi, talak, lcarlone\}@mit.edu}
}

\maketitle

\begin{abstract}
Semantic 3D scene understanding is a problem of critical importance in robotics. While significant advances have been made in simultaneous localization and mapping algorithms, robots are still far from having the common sense knowledge about household objects and their locations of an average human. We introduce a novel method for leveraging common sense embedded within large language models for labelling rooms given the objects contained within. This algorithm has the added benefits of (i) requiring no task-specific pre-training (operating entirely in the zero-shot regime) and (ii) generalizing to arbitrary room and object labels, including previously-unseen ones -- both of which are highly desirable traits in robotic scene understanding algorithms. The proposed algorithm operates on 3D scene graphs produced by modern spatial perception systems, and we hope it will pave the way to more generalizable and scalable high-level 3D scene understanding for robotics.
\end{abstract}

\IEEEpeerreviewmaketitle

\section{Introduction}
\label{sec:introduction}

A key challenge in robotics is that of scene understanding. If robotic systems are to see widespread deployment, then they must be able to not only map and localize within a multitude of environments, but also have a semantic understanding of said environment and the entities within it. That is, if an autonomous agent is told to ``go to the kitchen and fetch a spoon,'' it should be able to (i) understand what a kitchen is and what objects are usually present within it, (ii) use that understanding to figure out what locations it has visited are likely to be kitchens, and (iii) segment out objects within the location to identify viable spoons to take.

These aspects are typically inferred using metric-semantic simultaneous localization and mapping (SLAM) algorithms, wherein a robotic agent must create a map of its environment, determine its location within said map, and annotate the map with semantic information on objects and locations. Modern spatial perception systems, like Kimera \citep{rosinol2021kimera} and Hydra \citep{hughes2022hydra}, arrange such semantic information in a 3D scene graph --  a data structure wherein nodes represent locations and entities, while edges represent paths or spatial relationships.

SLAM pipelines can compute geometric information (such as position and bounding box) for entities and regions in the scene. However, attaching semantic labels to these nodes still remains a major open obstacle, especially for nodes corresponding to high-level spatial concepts, like rooms and buildings. To label a room node, the system must consider what objects are in the room (e.g. if it contains a stove, sink, and refrigerator, the considered room is likely a kitchen). This therefore necessitates some ``common sense'' mechanism to provide the system with such knowledge. 

One largely unexplored candidate method for imparting this common sense is by using language models. As systems like word embeddings and language models are trained on large text corpora, they might capture some of the semantic information contained within said datasets. For instance, a language model like BERT \citep{devlin2019bert} may have learned that the sentence ``Bathrooms often contain \underline{\hspace*{.5cm}}.'' is better completed with the word ``toilets'' rather than ``stoves,'' thereby containing some of the common sense needed for scene understanding -- critically, even without any additional fine-tuning. 

Likewise, being what \citet{von1988language} and \citet{chomsky1965} famously called an ``infinite use of finite means,'' language provides a medium through which arbitrary common sense queries can be compactly made and evaluated, including ones containing novel concepts and entities. This is an attractive feature for spatial perception algorithms in robotics, since \textit{autonomous agents in general deployment will naturally come across a wide variety of unfamiliar objects -- including ones that the engineer did not expect during development}. Being able to still perform generalizable inference over these new and unfamiliar object types would thus be highly beneficial.

Previous papers like \citet{tellex2011nlcommands, sharma2022}, and more \citep{ languageconditionedil, cliport, Kollar2010TowardUN, tellex2014, Matuszek2013} have mainly leveraged language for communicating goals or instructions to a robot to plan around or execute. To our knowledge, natural language processing tools have not been used for robot room categorization. Past works like \citet{roomcat} instead use an explicit Bayesian probabilistic framework for determining room categories based on detected objects. However, these methods remain hard to generalize to new room and object categories, with \citet{roomcat} only considering five room labels. In hopes of addressing these shortcomings and capitalizing on the previously outlined benefits of natural language, in this project, we explore the ability for large language models to be used as common sense mechanisms for robot scene understanding.

\section{Contribution}
\label{sec:contribution}

\begin{figure}[h]
    \centering
    \includegraphics[width=.5\textwidth]{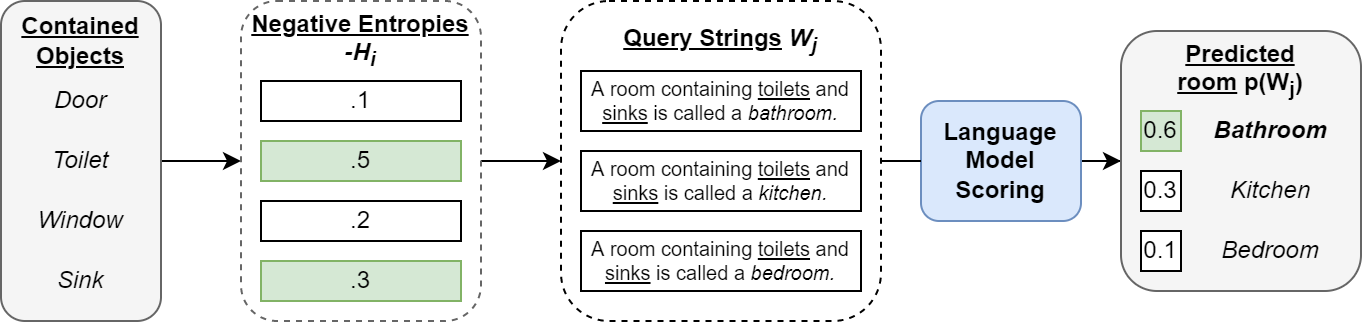}
    \caption{Example of experimental approach with $k = 2$ and $L_R = \{\text{Bathroom, Kitchen, Bedroom}\}$. The most informative object labels within the room are dynamically put into the query strings, which are then evaluated via language model.}
    \label{fig:example}
\end{figure}

We provide a novel framework for querying large language models to identify room labels in 3D scene graphs (e.g. as generated by Kimera \citep{rosinol2021kimera} or Hydra \citep{hughes2022hydra}) given the room's contained objects. Specifically, we use a heuristic method for picking out a small number of semantically-informative objects present within a room, constructing a query string with those object labels acting as a description of the room, and passing the string through a language model in order to infer the room's label (see Fig. \ref{fig:example}).

Because all inference over object and room labels occur by encoding the category names as strings, this method is able to handle arbitrary label input and output spaces. Additionally, we do not perform any training or fine-tuning, meaning all performance is entirely zero-shot.

\subsection{Language Model Evaluation}
\label{subsec:lm}
Language models are commonly broken into two types: (i) masked language models (MLMs) like BERT \citep{devlin2019bert} and RoBERTa \citep{roberta} and (ii) next-token prediction models, such as most GPT variants \citep{gpt-j, gpt-neo, GPT2}. Both are able to score an input sentence based off of semantic and grammatical sensibility. For MLMs, this can be done with pseudo-log likelihoods \citep{Salazar2020PLL}. For next token prediction models, which we focus on in this work, one can add up the log probabilities of each word/token in the sentence, given all previous ones, for the entire sentence. This effectively gives the log probability of the full sentence, as it is the factored probability distribution, as given by the (log) chain rule:
\begin{equation}
    \log p(W) = \log p([W]_1) + \sum_{i=2}^{|W|} \log p([W]_i\ |\ [W]_{1:i-1})
\end{equation}
where $W$ is the sentence and $[W]_i$ is the $i$-th word/token. Negating this summation gives the entropy (which is a common training objective for such models) and subsequently exponentiating gives the \textit{perplexity} of the sentence for the given model.

Returning to the example in Section \ref{sec:introduction}, we expect a sentence like ``Bathrooms often contain toilets.'' to be scored as being more likely than the sentence ``Bathrooms often contain stoves.'' We thus can use the scores of strings containing common sense facts as a proxy measure for how likely it is for the given fact to be true.

\subsection{Query Strings}
\label{subsec:querystrings}
We now require a method of creating a query string summarizing a given room whose label we wish to infer. To do this, we assume access to a list of objects within each room. On an actual robotic platform, this would be produced as part of its metric-semantic SLAM pipeline \citep{hughes2022hydra, rosinol2021kimera}. However, for our experiments, we use the ground truth object labels from our considered dataset (see Section \ref{subsec:dataset}).

Putting \textit{all} the objects in a room into the query may result in poor performance, as (i) long sentences are naturally less common and (ii) the queries may be dominated by objects that are present in many rooms (such as lights, doors, and windows). We thus wish to select only the $k$ most semantically informative objects. 

To do this, we first note that objects which only appear in a few room types are more informative, since their presence heavily implies certain room labels. Quantitatively, these objects have highly \textit{non}-uniform distributions $p(r_j \in L_R \ |\ o_i \in L_O)$, where $o_i$ is the object label, $r_j$ is the room label, and $L_{R,O}$ are the sets of room and object labels respectively. We compute these conditional probabilities in two ways:
\begin{enumerate}
    \item Using ground truth co-occurrency frequencies, i.e. finding how many times each object label appears in each type of room and normalizing over rooms. However, this does naturally necessitate some task-specific data. When using these empirical conditional probabilities, we also use Laplace smoothing.
    \item Using proxy co-occurrency probabilities by querying language models. Specifically, we use:
    \begin{equation}
        p(r_j \ |\ o_i) \approx \frac{\exp \log p(W_{i,j})}{\sum_{r_{j'} \in L_R} \exp\log p(W_{i,j'})}
    \end{equation}
    where $W_{i,j}$ is the query string ``A room containing $o_i$ is called a(n) $r_j$.'' and the overall log probability $\log p(W_{i,j})$ is computed using the method outlined in Section \ref{subsec:lm}. These conditional probabilities can be pre-computed for every room/object label pair.
\end{enumerate}
In either case, with $p(r_j \ |\ o_i)$ available, a natural measure of its non-uniformity is entropy:
\begin{equation}
    H_i = -\sum_{r_j \in L_R} p(r_j\ |\ o_i) \log p(r_j\ |\ o_i)
\end{equation}
Entropy is maximized when the considered distribution is uniform and minimized when it is a delta distribution, meaning more semantically-informative objects, which have less uniform distributions, have \textit{lower} corresponding $H_i$ values.

Therefore, in order to pick which labels to include in the query string, we just take the $k$ different lowest-entropy labels of present objects:
\begin{equation}
     O_{\text{best}} = \underset{o_i \in O}{\text{arg lowest\_$k$}} \ \left[ H_i \right]
\end{equation}
where $O$ is the set of all object labels contained within the considered room.

Finally, for a given room, we construct $|L_R|$ query strings, one per room label:
\begin{equation}
    \begin{split}
        W_j =\ &\text{``A room containing $o_1$, $o_2$... and $o_k$}\\
        &\text{is called a(n) $r_j$.''}       
    \end{split}
\end{equation}
$\forall r_j \in L_R $ and where $o_{1...k} \in O_{\text{best}}$, ordered by ascending entropy.

\subsection{Evaluation}
Lastly, to evaluate all the strings and make the final room label inference, we pass all the query strings $W_j,\ \forall r_j \in L_R$ into a language model to produce a set of scores:
\begin{equation}
    S(W_j) = \log p(W_j)
\end{equation}
The final inferred label is thus just the highest-scoring one, $\argmax_{r_j} S(W_j)$.

\section{Experiments}
\label{sec:experiments}
\subsection{Dataset}
\label{subsec:dataset}
We evaluate the above zero-shot algorithm on a scene graph dataset produced from the Matterport3D dataset \citep{Matterport3D}, which is commonly used in robot semantic navigation tasks \citep{habitat19iccv, habitatchallenge2022, szot2021habitat}. The Matterport3D dataset contains regions (rooms) and the objects contained within each room. Each region and object also has a semantic label and bounding box.

To convert these semantic meshes into scene graphs, we create a node for each region and object. Then, we connect all object nodes assigned to a region to that region's room node. Additional scene graph creation details are outlined in the Appendix.

We also filter out some regions. While Matterport3D contains outdoor regions as well (``yard,'' ``balcony,'' and ``porch''), we do not perform inference over them, since they are not true rooms and thus would require an alternate query string structure. In addition to outdoor regions, we also remove all rooms with no objects within or with the label of ``none.'' In total, after these filters, there are 1878 rooms.

Each object is assigned labels from several label spaces. We consider the original labels used by Matterport3D (mpcat40) and the labels used by NYU (nyuClass) \citep{nyuv2}. For both, we filter out nodes belonging to the mpcat40 categories ``ceiling,'' ``wall,'' ``floor,'' ``miscellaneous,'' ``object,'' and any other unlabeled nodes. We remove these categories because they are either not objects within the room or they are ambiguous to the point of being semantically uninformative. However, for nyuClass, we do \textit{not} reject objects classified by mpcat40 as ``object,'' since nyuClass has many more fine-grained and semantically-rich categories which all are mapped to this category. After pre-processing the label spaces in this way, mpcat40 has 35 object labels and nyuClass has 201. Both datasets share a room label space with 23 labels. See Table \ref{tab:labelfreqs} for a breakdown of room label frequencies.

\begin{table}[]
\centering
\caption{Room Label Frequencies in Pre-processed Dataset}
\label{tab:labelfreqs}
\resizebox{\columnwidth}{!}{%
\begin{tabular}{@{}ccccccc@{}}
\toprule
\textbf{Room Label} &
  \textit{Bar} &
  \textit{Bathroom} &
  \textit{Bedroom} &
  \textit{Classroom} &
  \textit{Closet} &
  \textit{\begin{tabular}[c]{@{}c@{}}Conference \\ Auditorium\end{tabular}} \\ \midrule
\textbf{Occurrences} &
  3 &
  365 &
  251 &
  2 &
  99 &
  16 \\
\textbf{Percentage} &
  0.16\% &
  19.43\% &
  13.37\% &
  0.11\% &
  5.27\% &
  0.85\% \\ \midrule
\textbf{Room Label} &
  \textit{\begin{tabular}[c]{@{}c@{}}Dining \\ Room\end{tabular}} &
  \textit{\begin{tabular}[c]{@{}c@{}}Family \\ Room\end{tabular}} &
  \textit{Game Room} &
  \textit{Garage} &
  \textit{Gym} &
  \textit{Hallway} \\ \midrule
\textbf{Occurrences} &
  74 &
  61 &
  17 &
  14 &
  16 &
  326 \\
\textbf{Percentage} &
  3.94\% &
  3.25\% &
  0.91\% &
  0.75\% &
  0.85\% &
  17.36\% \\ \midrule
\textbf{Room Label} &
  \textit{Kitchen} &
  \textit{\begin{tabular}[c]{@{}c@{}}Laundry \\ Room\end{tabular}} &
  \textit{Library} &
  \textit{\begin{tabular}[c]{@{}c@{}}Living \\ Room\end{tabular}} &
  \textit{Lobby} &
  \textit{Lounge} \\ \midrule
\textbf{Occurrences} &
  78 &
  35 &
  1 &
  71 &
  62 &
  64 \\
\textbf{Percentage} &
  4.15\% &
  1.86\% &
  0.05\% &
  3.78\% &
  3.30\% &
  3.41\% \\ \midrule
\textbf{Room Label} &
  \textit{Office} &
  \textit{Spa} &
  \textit{Staircase} &
  \textit{\begin{tabular}[c]{@{}c@{}}Television \\ Room\end{tabular}} &
  \textit{\begin{tabular}[c]{@{}c@{}}Utility \\ Room\end{tabular}} &
  \textit{\textbf{Total}} \\ \midrule
\textbf{Occurrences} &
  98 &
  44 &
  152 &
  13 &
  16 &
  \textbf{1878} \\
\textbf{Percentage} &
  5.22\% &
  2.34\% &
  8.09\% &
  0.69\% &
  0.85\% &
  \textbf{100\%} \\ \bottomrule
\end{tabular}%
}
\end{table}

\subsection{Trial Specifications}
We vary both whether we are using ground truth or proxy object/room label co-occurrencies (see Section \ref{subsec:querystrings}) and whether we are using the mpcat40 or nyuClass object label spaces, for a total of four trial conditions. We test on all rooms in our dataset, choosing $k=3$ objects per room to create the corresponding query sentences. For rooms containing fewer than three objects, we simply include as many as we can in the query string.

For all trials, we use the GPT-J next-token prediction language model, the largest open-source GPT-style model currently available \citep{gpt-j}. This model was used for both evaluating query strings and generating proxy co-occurrencies. Due to hardware limitations, we use the half-precision GPU release of the model. All experiments are run in PyTorch 1.8.0 \citep{pytorch}.

\subsection{Results}

\begin{table}[]
\centering
\caption{Inference Accuracies for All Four Conditions}
\label{tab:accs}
\resizebox{.6\columnwidth}{!}{%
\begin{tabular}{@{}ccc@{}}
\toprule
\textbf{}                                                                        & \textbf{nyuClass} & \textbf{mpcat40} \\ \midrule
\textbf{\begin{tabular}[c]{@{}c@{}}Ground Truth \\ Co-occurrencies\end{tabular}} & 52.41\%           & 49.36\%          \\ \midrule
\textbf{\begin{tabular}[c]{@{}c@{}}Proxy \\ Co-occurrencies\end{tabular}}        & 28.14\%           & 27.00\%          \\ \bottomrule
\end{tabular}%
}
\end{table}

\begin{figure*}[h]
    \centering
    \includegraphics[width=.95\textwidth]{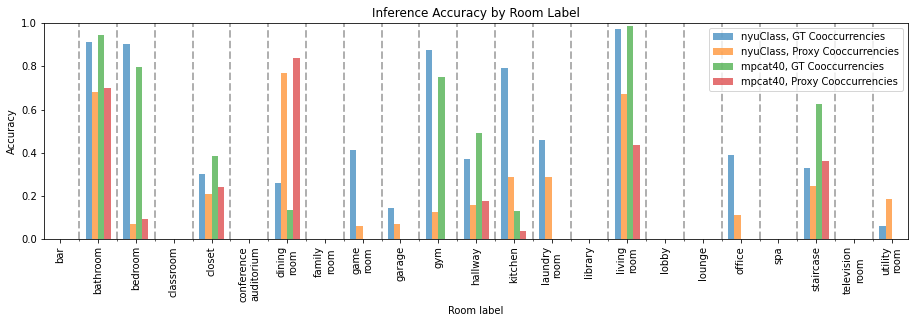}
    \caption{Inference accuracies of all four conditions, broken down by room label. Generally, ground truth co-occurrencies perform better than the proxy method. Additionally, note that nyuClass has higher performance in several room labels (e.g. kitchen) due to its more fine-grained object labels providing additional helpful information for inference.}
    \label{fig:cataccs}
\end{figure*}

Table \ref{tab:accs} shows that all four of our trials showed relatively high inference accuracies ($27 - 52.41\%$), scoring higher than both random chance (uniform over rooms, so $4.35\%$) and a naive baseline of always choosing the most frequent room label (bathroom, which is $19.44\%$ of the rooms in our dataset). 

Looking at Fig. \ref{fig:cataccs}, our algorithm achieves very high identification accuracies for several common household room types (``bathroom,'' ``bedroom,'' ``kitchen,'' and ``living room''). For the best-performing trial, nyuClass with ground truth co-occurrencies, accuracies for these key rooms range from $79.22 - 97.14\%$. Furthermore, there seem to be two general trends for when a room will \textit{not} be classified correctly:
\begin{enumerate}
    \item \textbf{If the room does not have disambiguating objects:} Bathrooms, staircases, and bedrooms all have good performance because they have objects that are almost exclusively found in them (e.g. showers/toilets, stairs/railing, and beds respectively). Rooms like conference auditoriums and family rooms, however, instead often just contain objects commonly found in many other rooms (tables, chairs, etc), and so are harder to identify.
    \item \textbf{If the room is not a ``standard'' room:} In our room label space $L_R$, we have ``bars,'' ``libraries,'' and ``spas,'' all of which more commmonly refer to \textit{buildings}, not rooms. Thus, query sentences $W_j$ for those rooms (e.g. ``A room containing ... is called a bar.'') are likely to be scored worse than ones with room labels that are unambiguously rooms. ``Gym'' should also fall into this category, but it has highly disambiguating object labels (e.g. ``gym equipment'' in mpcat40), so our algorithm identifies it correctly nonetheless.
\end{enumerate}
For the rest of the rooms, however, the language model does demonstrate the desired common sense when classifying them based off of objects typically found within each category of room.

Finally, the results demonstrate significant zero-shot generalization performance, as our model is able to handle both the smaller, 35-object label space (mpcat40) and the much larger, 201-object label space (nyuClass) as well. In fact, the nyuClass trials both resulted in higher accuracies than their mpcat40 counterparts (see Table \ref{tab:accs}). This is because nyuClass's labels are more specific and, therefore, more semantically-informative.

This benefit is best demonstrated when identifying four room labels: kitchens, laundry rooms, game rooms, and garages.
\begin{enumerate}
    \item \textbf{Kitchens and laundry rooms:} Both these rooms are characterized by the appliances they contain. For nyuClass, objects like stoves and refrigerators are almost exclusively found in the former while washing machines are found in the latter. However, mpcat40 groups all those objects under the very broad and ambiguous category of ``appliances,'' making differentiation between the two room labels more difficult.
    \item \textbf{Game rooms:} Game rooms are characterized by recreational objects, such as ping-pong tables and foosball tables. Both of these categories appear in nyuClass, but are simply classified as ``tables'' by mpcat40, again making it easier to identify when using the larger label space.
    \item \textbf{Garages:} Similar to the previous case, garage doors (which appear in nyuClass) are classified as just ``doors'' in mpcat40, again making garages easier to identify when using the larger label space.
\end{enumerate}

\section{Conclusion} 
\label{sec:conclusion}
We show how large language models can be used for robot scene understanding by introducing a novel algorithm for inferring room labels given its contained objects in 3D scene graphs. We show that this procedure is able to achieve good inference accuracy even in the zero-shot regime. Moreover, we also demonstrate that, not only can it \textit{handle} different label spaces, both large and small, it actively \textit{benefits} from the added semantic informativeness of more fine-grained and specific object labels. These traits make it a promising avenue of development for scalable, sample-efficient, and generalizable robot spatial perception systems.

\section*{Acknowledgments}
\noindent
We thank Prof. Jacob Andreas for his valuable feedback, discussion inputs, and suggestions. This work was partially funded by ARL DCIST CRA W911NF-17-2-0181 and ONR RAIDER N00014-18-1-2828.

\bibliographystyle{plainnat}
\bibliography{references}

\appendix

We perform a few additional dataset pre-processing steps to produce the final scene graph dataset with 1878 rooms. First, since some objects are assigned an incorrect region (e.g. toilets are assigned to living rooms, despite (i) that being non-sensible and (ii) the toilet not being within the bounding box of the living room), we check to see if each object is within the bounding box of its assigned region. If not, then it is re-assigned to whichever region's bounding box contains it, and the corresponding scene-graph connection is also made. 

Second, nyuClass has some misspelled labels (e.g. ``refridgerator'' instead of ``refrigerator''), so we correct all of those too. 

Lastly, sometimes, a single nyuClass label may be erroneously assigned to multiple mpcat40 labels. This is most problematic when one of the mpcat40 labels is rejected and the other is not. To address this, we use the first mpcat40 label for each nyuClass label that is \textit{not} rejected (e.g. nyuClass label ``stairs'' is mapped to mpcat40 ``miscellaneous,'' which is rejected, and ``stairs,'' which is not, so we keep the latter). However, this means some labels which \textit{should} be rejected are not rejected, so we also manually filter out all nyuClass object labels that are the \textit{same} as those of rejected mpcat40 labels: ``ceiling,'' ``floor,'' and ``wall''.

\end{document}